\documentclass[11pt]{article}

\usepackage[final]{acl}

\usepackage{times}
\usepackage{latexsym}

\usepackage[T1]{fontenc}
\usepackage[utf8]{inputenc}
\usepackage{amsmath}
\usepackage{amsfonts}
\usepackage{booktabs}
\usepackage{tabularx}
\usepackage{microtype}
\usepackage{multirow}

\usepackage{inconsolata}

\usepackage{graphicx}
\usepackage{adjustbox}
\usepackage{enumitem}
\usepackage{xspace}

\newcommand{\score}{\textsc{TAIScore}\xspace}

\title{Co-Evolving Actor-Conditioned Critics for Non-Verifiable Generation}
\author{
\textbf{Jinyoung Kim}$^1$ \quad 
\textbf{Muhammad Khalifa}$^1$ \quad 
\textbf{Lajanugen Logeswaran}$^2$ \quad \\
\textbf{Jaekyeom Kim}$^2$ \quad 
\textbf{Moontae Lee}$^{2,3}$ \quad 
\textbf{Honglak Lee}$^{1,2}$ \quad 
\textbf{Lu Wang}$^1$ \\
\\
$^1$University of Michigan \quad $^2$LG AI Research \quad $^3$University of Illinois at Chicago
}

\begin{document}
\maketitle
\begingroup
\renewcommand{\thefootnote}{\fnsymbol{footnote}}
\footnotetext{Correspondence to \texttt{jinykim@umich.edu}.}
\endgroup
\begin{abstract}
Natural-language critiques provide supervision beyond scalar rewards for non-verifiable generation, which lacks deterministic verifiers. In critique-guided refinement, a critic gives feedback on an initial response and an actor revises it.
However, final revision quality does not reveal whether the critique was actually useful: a capable actor may improve without following the feedback, while valid feedback may fail if the actor cannot execute it.
We frame critique as actor-conditioned revision guidance, where usefulness depends on whether the feedback helps the target actor address the intended weakness.
We introduce \textbf{\score} (\textbf{T}argeted \textbf{A}ctionable \textbf{I}mprovement \textbf{Score}), a reward that evaluates the instruction, initial response, critique, and revision together, assessing whether the critique targets a real weakness, whether the actor follows it, and whether the intended aspect improves.
We use this reward to train an actor-tailored critic with GRPO, and use critique-guided refinements to construct DPO preference pairs for the actor, forming a co-evolving critic-actor loop where the critic adapts to the actor's changing capability. 
Experiments show that an 8B critic trained with \score outperforms both a zero-shot 120B critic and critics trained with outcome-only or critique-only reward signals. Co-evolving the critic and actor further improves performance, suggesting that effective critique supervision should adapt as the actor changes.\footnote{
Our implementation can be accessed at \url{https://github.com/jiny1623/taiscore}.
}

\end{abstract}
\section{Introduction}
\label{sec:introduction}

Reinforcement learning from verifiable rewards
(RLVR)~\citep{shao2024deepseekmath, guo2025deepseek} has driven rapid progress on reasoning tasks: in mathematics or code generation, outputs can be checked against an answer key or unit test, giving learning algorithms a clear signal of task success~\citep{cobbe2021training, lightman2024let, le2022coderl}. However, extending this paradigm to open-ended tasks is difficult because such tasks admit no verifiable measure of success. In creative writing, complex instruction following, and long-form research generation, multiple valid responses may satisfy the same prompt, and quality is inherently multi-dimensional: a response may be fluent but incomplete, creative but unfaithful, or well-grounded but poorly organized~\citep{zhong2022towards, liang2022holistic}. Scalar preferences or judge scores can rank responses but cannot explain which dimension failed or what edit would help~\citep{wu2023fine, luo2025language}.

This has motivated growing interest in natural-language critique as a richer form of supervision~\citep{madaan2023self, shinn2023reflexion}.
A typical critique-guided revision pipeline involves two roles: a \emph{critic}, which provides feedback on an initial response, and an \emph{actor}, which revises the response based on that feedback.
Unlike a scalar score, a critique can identify \emph{which} aspect of a response is weak, explain \emph{why} it falls short, and propose to the actor model \emph{how} to fix it. A growing body of work uses such feedback to build critique-refinement pipelines that revise the actor model's outputs~\citep{scheurer2023training, wadhwa-etal-2024-learning-refine, yu2025training}. This reflects an important shift: improving non-verifiable generation requires not only judging outputs, but also providing actionable guidance for revision. However, existing approaches typically evaluate critiques either by their standalone quality or by the final improvement they coincide with, leaving open whether the target actor actually takes up the feedback and improves the intended aspect.

We argue that critique usefulness is determined not by the critique alone, but by its interaction with the actor that must execute it.
In our controlled analyses (\S\ref{sec:motivating-analysis}), we find that the same feedback can lead to substantially different revision behaviors and improvement gains depending on the actor’s capability. Conversely, increasing critic scale does not reliably produce more useful feedback for a fixed actor. 
These findings shift the critic-training objective: rather than rewarding critiques that look good in isolation or merely coincide with better final revisions, we aim to reward critiques that help the target actor make the intended change.

To operationalize this objective, we introduce 
\textbf{\score} (\textbf{T}argeted \textbf{A}ctionable \textbf{I}mprovement \textbf{Score}). Rather than scoring a critique by the final response quality or by its standalone plausibility, we evaluate the full critique-guided revision process
: the process in which an
actor revises an initial response to an instruction after receiving a critique.
A \score judge observes the instruction, initial response, critique, and revision together, and assesses whether the critique targets a real weakness, whether the actor follows it, and whether the intended aspect improves. We use this score as the critic-training reward. Critique-guided refinements are then used to construct preference pairs for actor DPO training, forming a co-evolving loop where the critic adapts as the actor improves.

Experiments on creative writing and deep research tasks show that an 8B critic
trained with \score outperforms both a stronger frozen \texttt{gpt-oss-120B}
critic and critics trained with only refinement quality or only critique
quality. Co-evolution further improves over the static \score critic, improving
the \texttt{Qwen3-8B} actor from 72.33 to 76.72 on WritingBench~\citep{wu2026writingbench}. The same trend holds on
HelloBench~\citep{que2024hellobench} and DeepResearch-Gym~\citep{coelho2025deepresearchgym}, suggesting that effective critique supervision
should be trained for the target actor rather than rely only on critic scale.

Our contributions are:
\begin{enumerate}
    \item We frame critique as actor-conditioned revision guidance and show that its usefulness depends on the critique-actor interaction rather than the critique alone.
    \item We propose \score, a reward that assigns credit to critiques by evaluating whether they identify valid issues, are followed by the actor, and lead to targeted improvement.
    \item We introduce a co-evolving critic-actor training procedure and
    demonstrate that critic training with \score outperforms
    training with only refinement quality or only critique quality, and that co-evolution outperforms static critics.
\end{enumerate}
\section{Related Work}
\label{sec:related_works}

\paragraph{Critique-guided self-refinement.}
A growing line of work uses natural-language feedback to improve model outputs through iterative revision. Self-Refine~\citep{madaan2023self} showed that a single model can generate, critique, and revise its own outputs in a test-time loop, while Reflexion~\citep{shinn2023reflexion} stores verbal feedback as memory across trials. 
Subsequent work has used language feedback as a training signal. 
For example, \citet{scheurer2023training} train models from natural-language
feedback at scale, while \citet{luo2025language} show that models can benefit
from verbal feedback without reducing it to scalar rewards.
More structured refinement pipelines decompose revision into separate
detect, critique, and refine stages~\citep{wadhwa-etal-2024-learning-refine}, synthesize reflection--revision trajectories for open-ended writing~\citep{liu2026r2}, or optimize critics with refinement-oriented signals~\citep{yu2025training}.
These methods demonstrate that critique can guide generation, but they
typically evaluate feedback through standalone critique plausibility or the
quality of the resulting revision. In contrast, we
evaluate whether a critique is useful for a specific actor by considering the
full critique-guided revision process it induces, including whether the actor
takes up the feedback and improves the intended aspect.

\paragraph{Reward modeling for non-verifiable generation.}
When tasks lack deterministic verifiers, reward signals often come from
learned or prompted evaluators.
RLHF~\citep{ouyang2022training} established the standard pipeline of
training reward models from human preferences, and
LLM-as-a-Judge~\citep{zheng2023judging} showed that strong LLMs can
approximate human judgments on open-ended tasks. Rubrics as
Rewards~\citep{gunjal2025rubrics} and
OpenRubrics~\citep{liu2025openrubrics} structure evaluation into
instance-specific criteria, improving interpretability and enabling
rubric-based RL for non-verifiable domains. However, these
criteria are typically aggregated into scalar signals for evaluating or
optimizing final responses. Such signals can rank responses or identify weak
dimensions, but they do not directly provide concrete revision guidance. In
contrast, natural-language critiques can specify what should be changed, why it
matters, and how the actor should revise the response.
 
\paragraph{Dynamic critics and actor-conditioned supervision.}
Static critics can become misaligned as the actor improves during training.
DR-Tulu~\citep{shao2025dr} addresses this by maintaining a rubric buffer that
co-evolves with the policy, %, generating new rubric items from on-policy rollouts.
RLAC~\citep{wu2025rlac} introduces an adversarial critic that proposes likely
failure modes for external verification. Recent work also connects critic training to actor
refinement behavior. CGI~\citep{yang2026lighthouse} jointly trains actors and critics
in interactive agent environments, where feedback helps actors refine actions
toward task success. Critique-RL~\citep{xi2025critique} trains critique
models with a two-stage RL procedure, combining direct rewards for critique
discriminability with indirect rewards from actor refinement.
These works demonstrate the value of adapting supervision to the current policy
or using actor refinement outcomes to train critics. Our work differs in both
setting and credit assignment. We study non-verifiable generation, where
rule-based correctness rewards are unavailable, and train the critic with a
judge-based reward that explicitly evaluates whether the critique targets a real
weakness, is taken up by the target actor, and improves the intended aspect.

\section{Motivation and Problem Formulation}
\label{sec:motivation}

Critique-guided revision consists of an instruction $x$, an initial
response $y_0$, a critique $c$, and a revised response $y_1$. In this
process, a critique is useful not simply when it appears valid, specific,
or actionable in isolation, but when it helps a specific actor revise the
response in the intended way. Feedback may correctly identify a weakness
yet be too abstract or too difficult for the current actor to execute;
conversely, feedback may be easy to follow but induce only superficial
edits or target low-priority issues. A useful critique should therefore be
both executable for the actor and substantive enough to produce an
attributable improvement in the targeted aspect of the response.

\subsection{Motivating Analysis: Critique Usefulness Is Actor-Conditioned}
\label{sec:motivating-analysis}

\begin{figure}[t]
    \centering
    \includegraphics[width=\columnwidth]{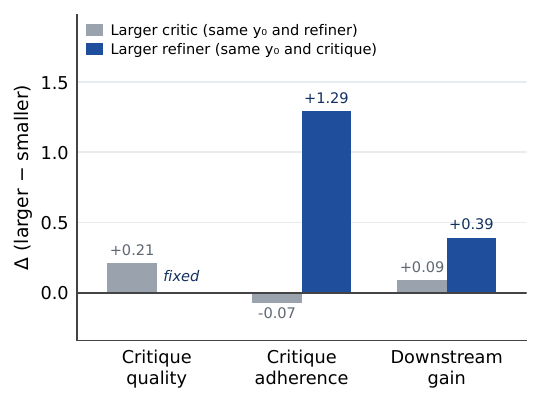}
    \caption{Controlled zero-shot analysis of critique-guided refinement on WritingBench. Bars show mean differences between larger- and smaller-model conditions in critique quality, critique adherence (how well $y_1$ implements $c$), and response-quality gain $S(y_1)-S(y_0)$. Gray bars compare larger and smaller critics while holding $y_0$ and the refiner fixed; blue bars compare larger and smaller refiners while holding $y_0$ and the critique fixed. Larger critics improve standalone critique quality but yield little average improvement in adherence or gain, whereas larger refiners substantially improve both. Individual comparisons are reported in Table~\ref{tab:motivation-details}.}
    \label{fig:motivation}
\end{figure}

To test whether critique usefulness is intrinsic to the critique or
conditioned on the actor, we conduct two controlled analyses on
WritingBench~\citep{wu2026writingbench}, varying one role at a time. The
critic-scaling comparison holds the instruction, initial response, and
revising actor fixed while replacing a smaller critic with a larger one;
the refiner-scaling comparison holds the instruction, initial response, and
critique fixed while replacing a smaller refiner with a larger one.

For each rollout $(x,y_0,c,y_1)$ we report three quantities.
\emph{Critique quality} is a prompted \texttt{gpt-oss-120B} judge score
rating whether the critique is valid, specific, important, actionable, and
faithful; it observes only $(x,y_0,c)$ and never sees the revision.
\emph{Critique adherence} is a second judge score that observes
$(x,y_0,c,y_1)$ and rates whether the actor incorporates the critique
rather than making unrelated edits. \emph{Gain} is the downstream
task-quality improvement $S(y_1)-S(y_0)$ under the WritingBench evaluator.
Judge prompts and implementation details are given
in Appendix~\ref{app:zero_shot_details}.

Figure~\ref{fig:motivation} summarizes these comparisons, with individual
comparison-level deltas reported in Appendix
(Table~\ref{tab:motivation-details}).

\paragraph{Same actor, different critics.}
If larger critics always produced more useful feedback for a given actor,
replacing a smaller critic with a larger one should improve not only
standalone critique quality, but also critique adherence and refinement gain.
Figure~\ref{fig:motivation} shows that this is not the case. Larger critics consistently receive higher standalone critique-quality
scores, but this improvement does not reliably translate into stronger critique adherence or
larger downstream gains. In some cases, the same actor incorporates the larger
critic's feedback less faithfully (Table~\ref{tab:motivation-details}). This suggests that critique quality judged
in isolation is not sufficient to predict usefulness for a fixed actor. Paired examples in Appendix~\ref{app:critic-reachability} illustrate one underlying
mechanism.

\paragraph{Same critique, different actors.}
If critique usefulness were determined by the critique alone, different
actors should benefit similarly from the same feedback. Instead,
Figure~\ref{fig:motivation} shows that larger actors exhibit stronger
adherence and larger refinement gains given identical critiques. Critique
usefulness is therefore not an intrinsic property of the critique, but
depends on the actor that executes it.

\subsection{Actor-Conditioned Critic Training}
\label{sec:problem-formulation}

These results motivate training critics for a particular actor. We reward
feedback only when it identifies a real weakness, is followed by that
actor, and produces improvement on the targeted aspect. This avoids two
forms of misattribution: standalone critique quality ignores
executability, whereas final response quality does not establish that the
critique caused the gain.  The objective is also
dynamic: as the actor improves, the critic must remain aligned with its
current revision capability, providing feedback that is executable but
still pushes the actor toward better revisions. In \S\ref{sec:method}, we
operationalize this objective with \score.
\section{Method}
\label{sec:method}

\begin{figure*}[t]
    \centering
    \includegraphics[width=\textwidth]{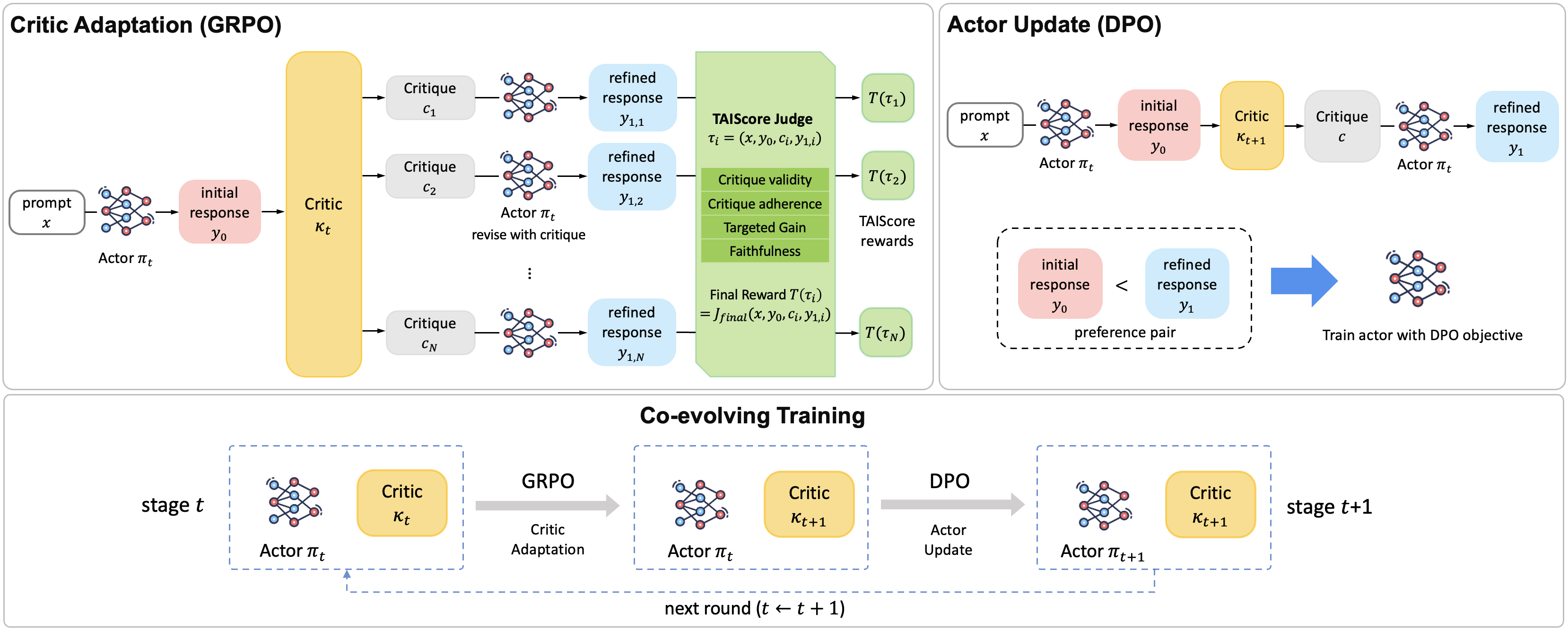}
    \caption{
Overview of our co-evolving critic-actor training framework.
For the current actor $\pi_t$, the critic $\kappa_t$ samples multiple critiques
for the same initial response $y_0$. Each critique produces a critique-guided
revision rollout $\tau_i=(x,y_0,c_i,y_{1,i})$. \score evaluates each rollout by
considering four diagnostic criteria: critique validity, critique adherence, targeted gain, and faithfulness, and then produces a final score $T(\tau_i)$
used as the GRPO reward for critic adaptation.
The actor $\pi_t$ then produces candidate revisions using critiques from the
adapted critic $\kappa_{t+1}$.
Revisions preferred over their corresponding
initial responses are selected to construct DPO preference pairs
$y_1 \succ y_0$ for updating the actor. Alternating these two updates yields
co-evolving critics and actors.
    }
    \label{fig:critique_conditioned_transition_generation}
\end{figure*}

Our goal is to train a critic whose feedback is useful as revision guidance
for a specific actor, and to keep the critic aligned as the actor improves.
We define \score, which assigns credit to a critique by examining the full
critique-guided revision process $(x,y_0,c,y_1)$~(\S\ref{sec:transition-scoring});
use it as a reward to update the critic via grouped policy
optimization~(\S\ref{sec:critic-update}); distill successful
critique-guided revisions into the actor through preference
optimization~(\S\ref{sec:actor-update}); and alternate the two updates so
that the critic tracks the actor's evolving
capability~(\S\ref{sec:coevolution}). Figure~\ref{fig:critique_conditioned_transition_generation} illustrates
the pipeline.
% \subsection{Critique-Revision Rollouts}

Throughout, we use the following notation. At training round $t$, $\pi_t$
denotes the current actor and $\kappa_t$ the current critic. For each instruction $x$, the actor generates an initial response
$y_0$, the critic generates a critique $c$, and the same actor generates a
critique-conditioned revision $y_1$.
We denote the resulting critique-guided revision rollout as
$\tau = (x, y_0, c, y_1)$.
% We use such transitions for two purposes: assigning reward to critiques and
% constructing preference data for actor training when the revision is judged to
% be a successful critique-guided improvement.

\subsection{Targeted Actionable Improvement Score}
\label{sec:transition-scoring}

A central design question is how to assign credit to a critique. Scoring
critique quality and revision quality separately risks incoherent supervision:
a generic or irrelevant critique may receive high reward if the revision happens
to improve for unrelated reasons, while a targeted critique may be
under-rewarded if the actor fails to execute it. To avoid such misattribution,
we define \score to evaluate a critique together with the revision it induces.
Given the full context $(x, y_0, c, y_1)$, \score assesses whether the critique
targets a real weakness, whether the actor follows it, and whether the targeted
aspect improves.

When computing \score, the judge first returns four diagnostic scores:
\[
(q_{\mathrm{qual}}, q_{\mathrm{adh}}, q_{\mathrm{gain}},
q_{\mathrm{faith}})
=
J_{\mathrm{diag}}(x,y_0,c,y_1).
\]
These correspond to the desiderata identified in
\S\ref{sec:motivating-analysis}: $q_{\mathrm{qual}}$ assesses \emph{critique validity}: whether the feedback is faithful to the response, specific,
important, and actionable; $q_{\mathrm{adh}}$ measures \emph{critique adherence}: whether
the actor incorporates the critique into the revision; $q_{\mathrm{gain}}$
captures \emph{targeted improvement}: whether the revision improves over the
initial response on the dimension the critique addresses; and
$q_{\mathrm{faith}}$ serves as a guardrail, verifying that the critique and
revision remain aligned with the original instruction.

Conditioned on these diagnostics, the judge then produces a final scalar:
\[
T(\tau) = J_{\mathrm{final}}(x, y_0, c, y_1) \in [1, 10].
\]
Because the diagnostic scores are generated before the final score within the
same inference pass, they provide an explicit reasoning scaffold for judging
the full critique-guided revision process. The diagnostic scores are used for
analysis, while the final score $T(\tau)$ serves as the critic training reward.

\subsection{Critic Update}
\label{sec:critic-update}
We update the critic using grouped critique-guided revision rollouts. At round
$t$, for each on-policy initial response $y_0 \sim \pi_t(\cdot \mid x)$, the
critic samples $N$ critiques $\{c_i\}_{i=1}^{N}$ conditioned on $(x, y_0)$.
The actor revises the same initial response under each critique, yielding
rollouts $\tau_i = (x, y_0, c_i, y_{1,i})$, each scored by \score as
$r_i = T(\tau_i)$.

Because all $N$ critiques are applied to the same initial response and revised
by the same actor, each group controls for the prompt, initial response, and
actor capability. The resulting rewards therefore provide relative credit for
which critique offers more useful revision guidance. We update the critic with
a group-relative policy-gradient (GRPO) objective~\citep{shao2024deepseekmath},
using normalized \score values as advantages:
\[
A_i
=
\frac{r_i-\mathrm{mean}(\{r_j\}_{j=1}^{N})}
{\mathrm{std}(\{r_j\}_{j=1}^{N})},
\]
% \vspace{-3mm}
\[
\mathcal{J}_t(\kappa)
=
\mathbb{E}
\left[
\frac{1}{N}
\sum_{i=1}^{N}
A_i
\log \kappa(c_i \mid x,y_0)
\right].
\]
This objective encourages the critic to place higher probability on critiques
that provide more useful revision guidance for the current actor $\pi_t$.

\subsection{Actor Update from Critique-Guided Refinements}
\label{sec:actor-update}
We use critique-guided revisions to update the actor via preference optimization.
Using the adapted critic $\kappa_{t+1}$, we generate critiques and corresponding
revisions from the current actor $\pi_t$. We construct preference pairs from
revisions that improve upon their corresponding initial responses under a pairwise assessment of response quality.
Each selected revision defines a preference pair $(y^+, y^-)=(y_1, y_0)$.

At each round, we construct a preference dataset of $M$ selected pairs:
\[
\mathcal{D}_t
=
\left\{
(x_i,y_i^+,y_i^-)
\right\}_{i=1}^{M},
\ \
(y_i^+,y_i^-)=(y_{1,i},y_{0,i}).
\]
We then update the actor using DPO. The actor is initialized from $\pi_t$, while
$\pi_{\mathrm{ref}}$ is a frozen copy of $\pi_t$ before the update:
\[
\begin{gathered}
\mathcal{L}_{\mathrm{DPO}}(\pi)
=
-\mathbb{E}_{(x,y^+,y^-)\sim\mathcal{D}_t}
\left[
\log \sigma
\left(
\beta \Delta_{\pi}
\right)
\right], \\
\Delta_{\pi}
=
\log \frac{\pi(y^+ \mid x)}
          {\pi_{\mathrm{ref}}(y^+ \mid x)}
-
\log \frac{\pi(y^- \mid x)}
          {\pi_{\mathrm{ref}}(y^- \mid x)} .
\end{gathered}
\]
This update encourages the actor to directly produce responses closer to those
it generates after receiving useful critique.

\subsection{Co-Evolving Critic and Actor}
\label{sec:coevolution}

A static critic can become misaligned as the actor changes during training.
The actor's failure modes change as it learns, and its ability to execute
feedback also changes. A critique that was useful for an earlier actor may
become redundant for a stronger actor, while feedback that was previously too
difficult may become well-applicable. We therefore alternate critic and actor updates across training rounds. Starting from $(\pi_0, \kappa_0)$, each
round~$t$ consists of two steps:
\begin{enumerate}
\item \textbf{Critic adaptation.} Fix the actor $\pi_t$ and update the critic
  via GRPO on grouped critique-guided revision rollouts:
  $\kappa_{t+1} \leftarrow \mathrm{GRPO}(\kappa_t; \mathcal{J}_t)$.
\item \textbf{Actor update.} Use $\kappa_{t+1}$ to generate
  critique-guided revisions from $\pi_t$, construct a preference dataset $\mathcal{D}_t$ from selected revisions,
  and update the actor:
  $\pi_{t+1} \leftarrow \mathrm{DPO}(\pi_t; \mathcal{D}_t)$.
\end{enumerate}
This yields a sequence $(\pi_0, \kappa_0), \ldots, (\pi_T, \kappa_T)$ in which
the critic continually adapts to the actor's current weaknesses and revision
capabilities, while the actor internalizes progressively stronger
critique-guided revisions.

\section{Experiments}
\label{sec:experiments}

\newcommand{\pmstd}[1]{{\scriptstyle \pm #1}}
\newcommand{\best}[1]{\mathbf{#1}}
\newcommand{\second}[1]{\underline{#1}}

\newcommand{\runstd}[1]{\ensuremath{{\scriptstyle \pm #1}}}

\begin{table*}[!t]

\centering

\small

\setlength{\tabcolsep}{4pt}

\begin{tabular}{lcccccc}

\toprule

\multirow{2}{*}{\textbf{Method}}
& \textbf{WritingBench}
& \multicolumn{2}{c}{\textbf{HelloBench}}
& \multicolumn{3}{c}{\textbf{DeepResearch-Gym}} \\

\cmidrule(lr){2-2}
\cmidrule(lr){3-4}
\cmidrule(lr){5-7}

& \textbf{Overall}($\uparrow$)
& \textbf{OEQA}($\uparrow$)
& \textbf{HTG}($\uparrow$)
& \textbf{KPR}($\uparrow$)
& \textbf{KPC}($\downarrow$)
& \textbf{Quality}($\uparrow$) \\

\midrule

\texttt{Qwen3-8B}
& 72.33\runstd{0.08}
& 34.86\runstd{1.95}
& 39.14\runstd{1.76}
& 71.93\runstd{0.04}
& 1.15\runstd{0.09}
& 81.89\runstd{0.07} \\

\midrule

\multicolumn{7}{l}{\textit{DPO pairs from off-the-shelf critics}} \\

\texttt{gpt-oss-120B} critic
& 75.41\runstd{0.14}
& 36.01\runstd{2.36}
& 50.93\runstd{2.18}
& 73.46\runstd{0.09}
& 1.11\runstd{0.08}
& 82.51\runstd{0.06} \\

\midrule

\multicolumn{7}{l}{\textit{DPO pairs from trained critics}} \\

Outcome-gain reward
& 75.63\runstd{0.09}
& 35.35\runstd{2.68}
& 45.14\runstd{2.48}
& 74.37\runstd{0.11}
& 1.12\runstd{0.09}
& 82.38\runstd{0.07} \\

Critique-quality reward
& 75.18\runstd{0.11}
& 35.51\runstd{2.45}
& 49.75\runstd{4.32}
& 74.19\runstd{0.13}
& 1.14\runstd{0.11}
& 82.32\runstd{0.04} \\

\score
& \underline{75.96}\runstd{0.12}
& \underline{36.48}\runstd{2.56}
& \underline{53.78}\runstd{1.58}
& \underline{75.21}\runstd{0.09}
& \underline{1.09}\runstd{0.07}
& \underline{82.63}\runstd{0.06} \\

\midrule

\multicolumn{7}{l}{\textit{Co-evolving critic-actor training}} \\

\score{} + co-evolution
& \textbf{76.72}\runstd{0.14}
& \textbf{39.84}\runstd{2.92}
& \textbf{54.62}\runstd{1.92}
& \textbf{76.14}\runstd{0.13}
& \textbf{1.03}\runstd{0.05}
& \textbf{83.15}\runstd{0.06} \\

\bottomrule

\end{tabular}

\caption{
Main results averaged over multiple runs, with standard deviations reported
after $\pm$. Except for the initial actor, all rows report the final
performance of the \texttt{Qwen3-8B} actor after DPO training.
Off-the-shelf critic baselines use untrained critics to construct revision
pairs, while trained-critic baselines first train the critic with different
reward definitions and then use the resulting critic to construct DPO pairs.
\score{} trains the critic with actor-conditioned supervision that rewards
valid, adopted, and targeted critiques, and co-evolving the critic with the
actor further updates the critic as the actor changes. Best results are shown
in bold and second-best results are underlined.
}
% \vspace{-2mm}
\label{tab:main-results}

\end{table*}

We design experiments to answer three questions:
\begin{enumerate}
    \item Does \score provide a better critic training signal
than outcome-only or critique-only rewards?
    \item Does co-evolving the critic and actor improve over a static critic?
    \item Do critiques tailored to one actor transfer to other actors?
\end{enumerate}

\subsection{Setup}
\label{sec:setup}

\paragraph{Tasks and data.}
We evaluate on two non-verifiable domains: creative writing and deep research.
For creative writing, we use DeepWriting-20K~\citep{wang2025reverse} as the training prompt source and
evaluate on WritingBench~\citep{wu2026writingbench} and HelloBench~\citep{que2024hellobench}. Following \citet{wang2025reverse},
we report HelloBench results on two key subsets: Open-Ended QA (OEQA), which evaluates detailed and nuanced answer generation, and Heuristic Text Generation (HTG), which evaluates creative reasoning and stylistic fidelity in
long-form text generation. For deep research, we use OpenScholar queries~\citep{asai2026synthesizing} as the training prompt source and evaluate on DeepResearch-Gym~\citep{coelho2025deepresearchgym}. Following the query
filtering procedure of \citet{shao2025dr}, we score each query with an LM judge and
retain high-quality queries for training. We use 6K training queries for each
domain. Actor-training pairs are constructed from larger shared candidate-query pools, as described in Appendix~\ref{app:pair-construction}. Details of the query selection procedure are provided in Appendix~\ref{app:query_selection}.

\paragraph{Evaluation.}
For WritingBench, we follow the official evaluation protocol and report the
overall score from the official evaluator.
For HelloBench, we use the OpenCompass implementation with the official
\texttt{GPT-4o-mini} judge configuration and report results on the Open-Ended
QA (OEQA) and Heuristic Text Generation (HTG) subsets. For DeepResearch-Gym, we
use the benchmark's report-level evaluation scripts with \texttt{GPT-4.1-mini}
as the judge, using the officially retrieved ClueWeb22 materials as context~\citep{coelho2025aligning}. We
report key-point recall (KPR), key-point contradiction (KPC), and report
quality. KPR is the percentage of reference key points supported by the
generated report, while KPC is the percentage contradicted by the report; higher
KPR and lower KPC are better. Report quality is the judge-rated mean over
report-level criteria, scaled to a 0-100 range. For each method, we use the
corresponding checkpoint to generate three independent output sets under
the same decoding configuration. We evaluate each output set separately and
report the mean and standard deviation across the three evaluation scores.

\paragraph{Models.}
We use models from the \texttt{Qwen3} family~\citep{yang2025qwen3} throughout
our experiments. The default actor and critic are both
\texttt{Qwen3-8B}, chosen as a mid-scale model capable of meaningful
self-refinement while leaving room for improvement. All actors perform on-policy self-refinement: the same model
generates the initial response and produces the revision after receiving
the critique. \score and all judge-based reward computations are performed by \texttt{gpt-oss-120B}. Except for the zero-shot off-the-shelf critic baseline described below, \texttt{gpt-oss-120B} is used only as a judge and is never updated or used as the actor.

\paragraph{Reward judge and cross-model agreement.}
\score is computed by \texttt{gpt-oss-120B} throughout critic training.
To check that the resulting signal is not an artifact of this particular
judge, we independently rescore 800 rollouts ($4\times200$ training
prompts) with \texttt{Claude Opus 4.8}, a model
from a different provider, using the identical rubric. The two judges
select the same top-ranked candidate for 89.5\% of prompts and agree on
77.9\% of non-tied within-prompt pairwise orderings, indicating that the
relative rankings used to form GRPO advantages are largely judge-invariant.
Protocol details are provided in Appendix~\ref{app:judge-validation}.

\paragraph{Training.}
We train the critic with GRPO~\citep{shao2024deepseekmath}. For each
on-policy initial response, the critic samples $N{=}4$ critiques, the actor
produces a revision for each critique, and \score provides the GRPO reward.
We train the actor with DPO~\citep{rafailov2023direct} ($\beta{=}0.1$).
Before each actor update, each method generates one candidate revision per query from a shared query pool. A critique-blind pairwise judge compares each revision with its corresponding initial response, and we uniformly sample 2K preferred revisions to form DPO pairs. Across methods, we hold fixed the candidate queries, number of candidate revisions, pair-selection procedure, and number of DPO pairs.

Co-evolution alternates the two updates for three rounds. In each round, we
update the critic against the current actor using 2K queries and then update
the actor using 2K pairs constructed with the adapted critic. Single-stage
methods instead train on the corresponding totals of 6K critic-training
queries and 6K actor-training pairs in one stage. Full hyperparameters and pair-construction details are provided in Appendices~\ref{app:hyperparams} and~\ref{app:pair-construction}.

\paragraph{Baselines.}
All conditions use the same DPO actor-training pipeline and differ only in how
the critiques used to construct DPO pairs are produced.
(1) \emph{Off-the-shelf critic}: To test whether critic training is necessary,
we compare against a stronger frozen critic: \texttt{gpt-oss-120B} prompted
directly to provide revision feedback. This baseline asks whether a strong
general-purpose critic can replace actor-conditioned critic training with
\score-based supervision.
(2) \emph{Trained critics}: We train critics with two ablated reward objectives.
The outcome-gain reward uses the same LM judge as a pairwise evaluator between
the initial response $y_0$ and the revised response $y_1$, rewarding critiques
that lead to higher-quality revisions. The critique-quality reward asks the LM
judge to rate the critique itself using the rubric-for-critiques prompt,
without considering whether the actor follows the feedback or improves the
response.
(3) \emph{Ours}: \score trains the critic with the full reward defined over
the critique-guided revision process, while \score + co-evolution additionally
alternates critic and actor updates across rounds. Further details on reward
computation and judge prompts are provided in Appendix~\ref{app:reward_prompts}.

\begin{table}[t]
\centering
\small
\setlength{\tabcolsep}{3.5pt}
\renewcommand{\arraystretch}{1.08}

\begin{tabularx}{\columnwidth}{
    @{}>{\raggedright\arraybackslash}X
    *{3}{>{\centering\arraybackslash}p{2em}}@{}
}
\toprule
Comparison ($A$ vs.\ $B$) & $A$ & Tie & $B$ \\
\midrule
\score{} vs.\ \texttt{Qwen3-8B}
    & \textbf{42} & 5 & 3 \\

\score{} vs.\ \texttt{gpt-oss-120B} critic
    & \textbf{35} & 7 & 7 \\

\score{} + co-evol. vs.\ \score{}
    & \textbf{36} & 8 & 5 \\
\bottomrule
\end{tabularx}

\caption{Majority-vote preferences on 50 WritingBench prompts.
Columns report preferences for condition $A$, ties, and preferences
for condition $B$. Rows 2--3 each omit one prompt without a majority
decision.}
\label{tab:human-eval}
\end{table}

\subsection{Main Results}
\label{sec:main-results}
Table~\ref{tab:main-results} shows three consistent trends. First,
off-the-shelf critique supervision improves the initial \texttt{Qwen3-8B}
actor, but is not sufficient: the frozen \texttt{gpt-oss-120B} critic is
outperformed by the \score-trained critic on all six
metrics. This indicates that critic scale alone does not determine downstream
usefulness. Second, \score outperforms both reward ablations.
Outcome-gain training rewards better final revisions but cannot determine
whether the improvement is attributable to the critique, while critique-quality
training rewards plausible feedback without checking whether the actor can
execute it. By evaluating the full critique-guided revision process,
\score provides a more effective critic-training signal.
Third, co-evolving the critic with the actor further improves performance
across all benchmarks, suggesting that critic feedback should adapt as the
actor's revision capability changes.

\paragraph{Human evaluation.}
We conduct a blind pairwise evaluation on 50 randomly sampled WritingBench
prompts (Table~\ref{tab:human-eval}). Human preferences corroborate the automatic results: \score is
preferred over both the base actor and the off-the-shelf critic baseline.
Co-evolution is also preferred over single-round \score, providing
independent evidence for its benefit. Full protocol details are provided in Appendix~\ref{app:human-eval}.

\subsection{Direct Critique Utility Before Actor Training}
\label{sec:direct-refinement}

\begin{table}[t]
\centering
\footnotesize
\setlength{\tabcolsep}{4pt}
\renewcommand{\arraystretch}{1.03}
\begin{tabular}{@{}lcc@{}}
\toprule
Condition & $S(y_1)$ & $\Delta_{\mathrm{WB}}$ \\
\midrule
\multicolumn{3}{c}{\textbf{(a) Critic source / objective}} \\[1pt]
Off-the-shelf \texttt{gpt-oss-120B}
    & 73.82 & +1.49 \\
Outcome-gain reward
    & 74.44 & +2.11 \\
Critique-quality reward
    & 74.23 & +1.90 \\
\score
    & \textbf{75.11} & \textbf{+2.78} \\
\addlinespace[2pt]
\multicolumn{3}{c}{\textbf{(b) Critique-content control}} \\[1pt]
No critique
    & 73.30 & +0.97 \\
Generic critique
    & 73.34 & +1.01 \\
Shuffled \score critique
    & 72.88 & +0.55 \\
Matched \score critique
    & \textbf{75.11} & \textbf{+2.78} \\
\bottomrule
\end{tabular}

\caption{Direct refinement on WritingBench before actor DPO.
All conditions share prompts, initial responses
($S(y_0){=}72.33$), refiner, decoding, and evaluator; only the
feedback varies. $\Delta_{\mathrm{WB}}{=}S(y_1)-S(y_0)$.
All revisions are included without preference filtering.}
\label{tab:direct-refinement}
\end{table}

Table~\ref{tab:direct-refinement} tests whether the feedback produced
by different critics leads to higher-quality immediate revisions,
without updating the actor or selecting revisions through the
preference-pair filter. We hold the initial response
$y_0$, \texttt{Qwen3-8B} reviser, revision prompt, decoding
configuration, and evaluator fixed, and vary only the supplied
feedback.

The \score-trained critic yields the highest mean revision score,
improving WritingBench from 72.33 to 75.11 ($+2.78$), ahead of the
off-the-shelf, outcome-gain, and critique-quality critics (Table~3a). The
advantage of \score{} is thus already observable at direct refinement time,
before any actor update.

Panel (b) tests whether this reflects generic second-pass revision rather
than critique content. Revising with no critique or a generic critique
yields only $+0.97$ and $+1.01$, and a \score{} critique sampled from a
different example yields $+0.55$ despite preserving the source and general
form of the feedback. The matched critique exceeds the strongest control by
1.77 points, indicating that the gain depends on instance-specific,
correctly matched critique content rather than the actor's general ability
to improve on a second pass. Further details are provided in
Appendix~\ref{app:critique-content-controls}.

\vspace{-0.3em}
\subsection{Actor--Critic Matching} \label{sec:actor-matching}

Different actors may respond differently to the same critique: feedback that one actor can readily apply may be less useful to another. We therefore examine whether a critic is more effective when tailored to the actor that will use its feedback. We fix the target actor to \texttt{Qwen3-8B} and train three critics, tailored respectively to \texttt{Llama-3.2-3B}, \texttt{Qwen3-4B}, and \texttt{Qwen3-8B}. All three critics start from the same \texttt{Qwen3-8B} checkpoint and use the same \score training procedure and budget; only the actor they are tailored to changes. We use each critic to construct DPO training pairs for the target actor, holding the queries and actor-training procedure fixed.

\begin{table}[t]
\centering
\small
\setlength{\tabcolsep}{6pt}
\renewcommand{\arraystretch}{1.0}
\begin{tabular}{@{}lcc@{}}
\toprule
Critic tailored to & WritingBench & Gain \\
\midrule
None (base actor)       & 72.33          & -- \\
\texttt{Llama-3.2-3B}      & 74.62          & +2.29 \\
\texttt{Qwen3-4B}                & 75.08          & +2.75 \\
\texttt{Qwen3-8B} (matched)      & \textbf{75.96} & \textbf{+3.63} \\
\bottomrule
\end{tabular} \caption{Effect of matching the critic to the target actor. The target actor is \texttt{Qwen3-8B} in every condition. All critics share the same initialization and training setup and differ only in the actor to which they are tailored.} \label{tab:actor-matching} \end{table}

As shown in Table~\ref{tab:actor-matching}, all three critics lead to substantial improvements over the base actor. However, the critic tailored directly to the target \texttt{Qwen3-8B} actor achieves the largest gain of 3.63 points, outperforming the \texttt{Qwen3-4B}-tailored critic by 0.88 points and the \texttt{Llama-3.2-3B}-tailored critic by 1.34 points. Thus, critique behavior learned with one actor can remain useful to other actors, but matching the critic to the target actor produces the greatest downstream improvement.

\section{Conclusion}

We presented an actor-conditioned view of critique-guided revision and
introduced \score, which rewards critiques that identify real
weaknesses, are followed by the target actor, and cause targeted
improvements. Across creative writing and deep research, \score
outperforms outcome-gain and critique-quality rewards, while adapting
the critic as the actor changes yields further gains. These results
support evaluating critique supervision through the revision behavior
it induces in the target actor.
\section*{Limitations}

This work evaluates actor-tailored critique training on two non-verifiable
generation domains: creative writing and deep research. While these settings
capture important open-ended generation challenges, broader evaluation on
additional domains such as dialogue, multimodal generation, and long-form
instruction following would further clarify the scope of the approach. Our main
training experiments also focus on Qwen-family models, although our controlled
zero-shot analysis includes both Qwen and Llama models and shows similar
actor-conditioned trends. Future work could therefore validate the full training
procedure across more model families and scales. Finally, our co-evolution
procedure uses a fixed number of rounds over disjoint data shards; adaptive
schedules that decide when to update the critic or actor, or continuous
co-training strategies that reuse feedback across rounds, are natural directions
for future work.
\section*{Ethical Considerations}
\label{sec:ethics}

Our experiments use publicly available models, benchmarks, and datasets. We use these resources in accordance with their respective licenses and terms of use and cite their original creators in the main text. We also conduct a small-scale human evaluation of anonymized model outputs, as described in Appendix~\ref{app:human-eval}. The evaluation collects only pairwise response preferences and does not solicit sensitive personal information. Three university students whose primary language is English served as annotators. Each independently evaluated anonymized response pairs presented in randomized order.

As with any method that improves open-ended text generation, outputs may
reflect biases or harmful patterns present in the pretraining data of the
underlying models. Because our approach improves critique-guided revision, it
could also improve the fluency, persuasiveness, or apparent helpfulness of
undesirable outputs if applied without appropriate safeguards. These risks are
not unique to our method, but they are relevant to downstream use of any system
that improves language model generation quality. We therefore encourage applying
appropriate safety filtering, monitoring, and domain-specific safeguards when
using critique-guided revision systems in practice.

Our use of existing artifacts is limited to research-oriented training and
evaluation, consistent with their intended use. Any artifacts produced by this
work are intended for research use only. During query selection, we
applied filtering to exclude unsafe examples and prompts that request private or
personally identifying information. Since we rely on publicly released datasets
and benchmarks, we do not attempt de-anonymization or collect new personal
information.

AI assistants were used for language polishing, proofreading, and formatting
suggestions. All content, claims, experiments, and conclusions were reviewed and
verified by the authors.
\section*{Acknowledgements}

This work is supported by LG AI Research.
\bibliography{custom}

\begin{thebibliography}{31}
\providecommand{\natexlab}[1]{#1}

\bibitem[{Asai et~al.(2026)Asai, He, Shao, Shi, Singh, Chang, Lo, Soldaini, Feldman, D’Arcy et~al.}]{asai2026synthesizing}
Akari Asai, Jacqueline He, Rulin Shao, Weijia Shi, Amanpreet Singh, Joseph~Chee Chang, Kyle Lo, Luca Soldaini, Sergey Feldman, Mike D’Arcy, et~al. 2026.
\newblock Synthesizing scientific literature with retrieval-augmented language models.
\newblock \emph{Nature}, pages 1--7.

\bibitem[{Cobbe et~al.(2021)Cobbe, Kosaraju, Bavarian, Chen, Jun, Kaiser, Plappert, Tworek, Hilton, Nakano et~al.}]{cobbe2021training}
Karl Cobbe, Vineet Kosaraju, Mohammad Bavarian, Mark Chen, Heewoo Jun, Lukasz Kaiser, Matthias Plappert, Jerry Tworek, Jacob Hilton, Reiichiro Nakano, et~al. 2021.
\newblock Training verifiers to solve math word problems.
\newblock \emph{arXiv preprint arXiv:2110.14168}.

\bibitem[{Coelho et~al.(2025{\natexlab{a}})Coelho, Martins, Magalh{\~a}es, and Xiong}]{coelho2025aligning}
Jo{\~a}o Coelho, Bruno Martins, Jo{\~a}o Magalh{\~a}es, and Chenyan Xiong. 2025{\natexlab{a}}.
\newblock Aligning web query generation with ranking objectives via direct preference optimization.
\newblock In \emph{Proceedings of the 48th International ACM SIGIR Conference on Research and Development in Information Retrieval}, pages 2982--2986.

\bibitem[{Coelho et~al.(2025{\natexlab{b}})Coelho, Ning, He, Mao, Paladugu, Setlur, Jin, Callan, Magalh{\~a}es, Martins et~al.}]{coelho2025deepresearchgym}
Jo{\~a}o Coelho, Jingjie Ning, Jingyuan He, Kangrui Mao, Abhijay Paladugu, Pranav Setlur, Jiahe Jin, Jamie Callan, Jo{\~a}o Magalh{\~a}es, Bruno Martins, et~al. 2025{\natexlab{b}}.
\newblock Deepresearchgym: A free, transparent, and reproducible evaluation sandbox for deep research.
\newblock \emph{arXiv preprint arXiv:2505.19253}.

\bibitem[{Gunjal et~al.(2025)Gunjal, Wang, Lau, Nath, He, Liu, and Hendryx}]{gunjal2025rubrics}
Anisha Gunjal, Anthony Wang, Elaine Lau, Vaskar Nath, Yunzhong He, Bing Liu, and Sean Hendryx. 2025.
\newblock Rubrics as rewards: Reinforcement learning beyond verifiable domains.
\newblock \emph{arXiv preprint arXiv:2507.17746}.

\bibitem[{Guo et~al.(2025)Guo, Yang, Zhang, Song, Wang, Zhu, Xu, Zhang, Ma, Bi et~al.}]{guo2025deepseek}
Daya Guo, Dejian Yang, Haowei Zhang, Junxiao Song, Peiyi Wang, Qihao Zhu, Runxin Xu, Ruoyu Zhang, Shirong Ma, Xiao Bi, et~al. 2025.
\newblock Deepseek-r1 incentivizes reasoning in llms through reinforcement learning.
\newblock \emph{Nature}, 645(8081):633--638.

\bibitem[{Le et~al.(2022)Le, Wang, Gotmare, Savarese, and Hoi}]{le2022coderl}
Hung Le, Yue Wang, Akhilesh~Deepak Gotmare, Silvio Savarese, and Steven Chu~Hong Hoi. 2022.
\newblock Coderl: Mastering code generation through pretrained models and deep reinforcement learning.
\newblock \emph{Advances in Neural Information Processing Systems}, 35:21314--21328.

\bibitem[{Liang et~al.(2022)Liang, Bommasani, Lee, Tsipras, Soylu, Yasunaga, Zhang, Narayanan, Wu, Kumar et~al.}]{liang2022holistic}
Percy Liang, Rishi Bommasani, Tony Lee, Dimitris Tsipras, Dilara Soylu, Michihiro Yasunaga, Yian Zhang, Deepak Narayanan, Yuhuai Wu, Ananya Kumar, et~al. 2022.
\newblock Holistic evaluation of language models.
\newblock \emph{arXiv preprint arXiv:2211.09110}.

\bibitem[{Lightman et~al.(2024)Lightman, Kosaraju, Burda, Edwards, Baker, Lee, Leike, Schulman, Sutskever, and Cobbe}]{lightman2024let}
Hunter Lightman, Vineet Kosaraju, Yuri Burda, Harrison Edwards, Bowen Baker, Teddy Lee, Jan Leike, John Schulman, Ilya Sutskever, and Karl Cobbe. 2024.
\newblock Let's verify step by step.
\newblock In \emph{International Conference on Learning Representations}, volume 2024, pages 39578--39601.

\bibitem[{Liu et~al.(2025)Liu, Xu, Yu, Hong, Yang, Zhao, and Wang}]{liu2025openrubrics}
Tianci Liu, Ran Xu, Tony Yu, Ilgee Hong, Carl Yang, Tuo Zhao, and Haoyu Wang. 2025.
\newblock Openrubrics: Towards scalable synthetic rubric generation for reward modeling and llm alignment.
\newblock \emph{arXiv preprint arXiv:2510.07743}.

\bibitem[{Liu et~al.(2026)Liu, Zhang, Li, Lai, Wu, Lei, and Yan}]{liu2026r2}
Wanlong Liu, Bo~Zhang, Chenliang Li, Shaopeng Lai, Yuning Wu, Xuanyu Lei, and Ming Yan. 2026.
\newblock R2-write: Reflection and revision for open-ended writing with deep reasoning.
\newblock \emph{arXiv preprint arXiv:2604.03004}.

\bibitem[{Luo et~al.(2025)Luo, Liu, Liu, Du, Lin, Chen, Lu, and Pang}]{luo2025language}
Renjie Luo, Zichen Liu, Xiangyan Liu, Chao Du, Min Lin, Wenhu Chen, Wei Lu, and Tianyu Pang. 2025.
\newblock Language models can learn from verbal feedback without scalar rewards.
\newblock \emph{arXiv preprint arXiv:2509.22638}.

\bibitem[{Madaan et~al.(2023)Madaan, Tandon, Gupta, Hallinan, Gao, Wiegreffe, Alon, Dziri, Prabhumoye, Yang et~al.}]{madaan2023self}
Aman Madaan, Niket Tandon, Prakhar Gupta, Skyler Hallinan, Luyu Gao, Sarah Wiegreffe, Uri Alon, Nouha Dziri, Shrimai Prabhumoye, Yiming Yang, et~al. 2023.
\newblock Self-refine: Iterative refinement with self-feedback.
\newblock \emph{Advances in neural information processing systems}, 36:46534--46594.

\bibitem[{Ouyang et~al.(2022)Ouyang, Wu, Jiang, Almeida, Wainwright, Mishkin, Zhang, Agarwal, Slama, Ray et~al.}]{ouyang2022training}
Long Ouyang, Jeffrey Wu, Xu~Jiang, Diogo Almeida, Carroll Wainwright, Pamela Mishkin, Chong Zhang, Sandhini Agarwal, Katarina Slama, Alex Ray, et~al. 2022.
\newblock Training language models to follow instructions with human feedback.
\newblock \emph{Advances in neural information processing systems}, 35:27730--27744.

\bibitem[{Que et~al.(2024)Que, Duan, He, Mou, Zhou, Liu, Rong, Wang, Yang, Zhang et~al.}]{que2024hellobench}
Haoran Que, Feiyu Duan, Liqun He, Yutao Mou, Wangchunshu Zhou, Jiaheng Liu, Wenge Rong, Zekun~Moore Wang, Jian Yang, Ge~Zhang, et~al. 2024.
\newblock Hellobench: Evaluating long text generation capabilities of large language models.
\newblock \emph{arXiv preprint arXiv:2409.16191}.

\bibitem[{Rafailov et~al.(2023)Rafailov, Sharma, Mitchell, Manning, Ermon, and Finn}]{rafailov2023direct}
Rafael Rafailov, Archit Sharma, Eric Mitchell, Christopher~D Manning, Stefano Ermon, and Chelsea Finn. 2023.
\newblock Direct preference optimization: Your language model is secretly a reward model.
\newblock \emph{Advances in neural information processing systems}, 36:53728--53741.

\bibitem[{Scheurer et~al.(2023)Scheurer, Campos, Korbak, Chan, Chen, Cho, and Perez}]{scheurer2023training}
J{\'e}r{\'e}my Scheurer, Jon~Ander Campos, Tomasz Korbak, Jun~Shern Chan, Angelica Chen, Kyunghyun Cho, and Ethan Perez. 2023.
\newblock Training language models with language feedback at scale.
\newblock \emph{arXiv preprint arXiv:2303.16755}.

\bibitem[{Shao et~al.(2025)Shao, Asai, Shen, Ivison, Kishore, Zhuo, Zhao, Park, Finlayson, Sontag et~al.}]{shao2025dr}
Rulin Shao, Akari Asai, Shannon~Zejiang Shen, Hamish Ivison, Varsha Kishore, Jingming Zhuo, Xinran Zhao, Molly Park, Samuel~G Finlayson, David Sontag, et~al. 2025.
\newblock Dr tulu: Reinforcement learning with evolving rubrics for deep research.
\newblock \emph{arXiv preprint arXiv:2511.19399}.

\bibitem[{Shao et~al.(2024)Shao, Wang, Zhu, Xu, Song, Bi, Zhang, Zhang, Li, Wu et~al.}]{shao2024deepseekmath}
Zhihong Shao, Peiyi Wang, Qihao Zhu, Runxin Xu, Junxiao Song, Xiao Bi, Haowei Zhang, Mingchuan Zhang, YK~Li, Yang Wu, et~al. 2024.
\newblock Deepseekmath: Pushing the limits of mathematical reasoning in open language models.
\newblock \emph{arXiv preprint arXiv:2402.03300}.

\bibitem[{Shinn et~al.(2023)Shinn, Cassano, Gopinath, Narasimhan, and Yao}]{shinn2023reflexion}
Noah Shinn, Federico Cassano, Ashwin Gopinath, Karthik Narasimhan, and Shunyu Yao. 2023.
\newblock Reflexion: Language agents with verbal reinforcement learning.
\newblock \emph{Advances in neural information processing systems}, 36:8634--8652.

\bibitem[{Wadhwa et~al.(2024)Wadhwa, Zhao, Li, and Durrett}]{wadhwa-etal-2024-learning-refine}
Manya Wadhwa, Xinyu Zhao, Junyi~Jessy Li, and Greg Durrett. 2024.
\newblock \href {https://doi.org/10.18653/v1/2024.findings-emnlp.716} {Learning to refine with fine-grained natural language feedback}.
\newblock In \emph{Findings of the Association for Computational Linguistics: EMNLP 2024}, pages 12281--12308, Miami, Florida, USA. Association for Computational Linguistics.

\bibitem[{Wang et~al.(2025)Wang, Que, Xu, Liu, Zhou, Feng, Zhong, Ye, Yang, Huang et~al.}]{wang2025reverse}
Haozhe Wang, Haoran Que, Qixin Xu, Minghao Liu, Wangchunshu Zhou, Jiazhan Feng, Wanjun Zhong, Wei Ye, Tong Yang, Wenhao Huang, et~al. 2025.
\newblock Reverse-engineered reasoning for open-ended generation.
\newblock \emph{arXiv preprint arXiv:2509.06160}.

\bibitem[{Wu et~al.(2025{\natexlab{a}})Wu, Zhang, Min, Levine, and Kumar}]{wu2025rlac}
Mian Wu, Gavin Zhang, Sewon Min, Sergey Levine, and Aviral Kumar. 2025{\natexlab{a}}.
\newblock Rlac: Reinforcement learning with adversarial critic for free-form generation tasks.
\newblock \emph{arXiv preprint arXiv:2511.01758}.

\bibitem[{Wu et~al.(2025{\natexlab{b}})Wu, Mei, Yan, Li, Lai, Ren, Wang, Zhang, Wu, Jin, and Huang}]{wu2026writingbench}
Yuning Wu, Jiahao Mei, Ming Yan, Chenliang Li, Shaopeng Lai, Yuran Ren, Zijia Wang, Ji~Zhang, Mengyue Wu, Qin Jin, and Fei Huang. 2025{\natexlab{b}}.
\newblock \href {https://doi.org/10.52202/085713-1737} {Writingbench: A comprehensive benchmark for generative writing}.
\newblock In \emph{Advances in Neural Information Processing Systems}, volume 38, Main Conference. Curran Associates, Inc.

\bibitem[{Wu et~al.(2023)Wu, Hu, Shi, Dziri, Suhr, Ammanabrolu, Smith, Ostendorf, and Hajishirzi}]{wu2023fine}
Zeqiu Wu, Yushi Hu, Weijia Shi, Nouha Dziri, Alane Suhr, Prithviraj Ammanabrolu, Noah~A Smith, Mari Ostendorf, and Hannaneh Hajishirzi. 2023.
\newblock Fine-grained human feedback gives better rewards for language model training.
\newblock \emph{Advances in Neural Information Processing Systems}, 36:59008--59033.

\bibitem[{Xi et~al.(2025)Xi, Huang, Guo, Hong, Yang, Fan, Li, Chen, Ye, Yuan et~al.}]{xi2025critique}
Zhiheng Xi, Jixuan Huang, Xin Guo, Boyang Hong, Dingwen Yang, Xiaoran Fan, Shuo Li, Zehui Chen, Junjie Ye, Siyu Yuan, et~al. 2025.
\newblock Critique-rl: Training language models for critiquing through two-stage reinforcement learning.
\newblock \emph{arXiv preprint arXiv:2510.24320}.

\bibitem[{Yang et~al.(2025{\natexlab{a}})Yang, Li, Yang, Zhang, Hui, Zheng, Yu, Gao, Huang, Lv et~al.}]{yang2025qwen3}
An~Yang, Anfeng Li, Baosong Yang, Beichen Zhang, Binyuan Hui, Bo~Zheng, Bowen Yu, Chang Gao, Chengen Huang, Chenxu Lv, et~al. 2025{\natexlab{a}}.
\newblock Qwen3 technical report.
\newblock \emph{arXiv preprint arXiv:2505.09388}.

\bibitem[{Yang et~al.(2025{\natexlab{b}})Yang, Ye, Li, Yuan, zhang, Tu, Li, and Yang}]{yang2026lighthouse}
Ruihan Yang, Fanghua Ye, Jian Li, Siyu Yuan, yikai zhang, Zhaopeng Tu, Xiaolong Li, and Deqing Yang. 2025{\natexlab{b}}.
\newblock \href {https://doi.org/10.52202/085713-5491} {The lighthouse of language: Enhancing llm agents via critique-guided improvement}.
\newblock In \emph{Advances in Neural Information Processing Systems}, volume 38, Main Conference, pages 164647--164678. Curran Associates, Inc.

\bibitem[{Yu et~al.(2025)Yu, Xiang, Yang, Ke, Wen, Wang, Cheng, Zhang, Mu, Sun et~al.}]{yu2025training}
Tianshu Yu, Chao Xiang, Mingchuan Yang, Pei Ke, Bosi Wen, Cunxiang Wang, Jiale Cheng, Li~Zhang, Xinyu Mu, Chuxiong Sun, et~al. 2025.
\newblock Training language model to critique for better refinement.
\newblock In \emph{Findings of the Association for Computational Linguistics: ACL 2025}, pages 26760--26804.

\bibitem[{Zheng et~al.(2023)Zheng, Chiang, Sheng, Zhuang, Wu, Zhuang, Lin, Li, Li, Xing et~al.}]{zheng2023judging}
Lianmin Zheng, Wei-Lin Chiang, Ying Sheng, Siyuan Zhuang, Zhanghao Wu, Yonghao Zhuang, Zi~Lin, Zhuohan Li, Dacheng Li, Eric Xing, et~al. 2023.
\newblock Judging llm-as-a-judge with mt-bench and chatbot arena.
\newblock \emph{Advances in neural information processing systems}, 36:46595--46623.

\bibitem[{Zhong et~al.(2022)Zhong, Liu, Yin, Mao, Jiao, Liu, Zhu, Ji, and Han}]{zhong2022towards}
Ming Zhong, Yang Liu, Da~Yin, Yuning Mao, Yizhu Jiao, Pengfei Liu, Chenguang Zhu, Heng Ji, and Jiawei Han. 2022.
\newblock Towards a unified multi-dimensional evaluator for text generation.
\newblock In \emph{Proceedings of the 2022 Conference on Empirical Methods in Natural Language Processing}, pages 2023--2038.

\end{thebibliography}

\newpage
\appendix
% \raggedbottom
\clearpage
\section{Details of the Zero-Shot Controlled Analysis}
\label{app:zero_shot_details}

\paragraph{Benchmark.}
We use the official WritingBench evaluation set for the zero-shot controlled
analysis. All controlled comparisons are evaluated on the same 1,000 examples. 

\paragraph{Critic-scaling comparison.}
We fix the instruction, initial response, and revising actor, and compare
feedback from a smaller critic and a larger critic. The same actor revises the
same initial response under each critique. This comparison isolates the effect
of critic scale.

\paragraph{Actor-scaling comparison.}
We fix the instruction, initial response, and critique, and compare revisions
from a smaller actor and a larger actor. This comparison isolates the effect of
actor capability.

\paragraph{Metrics.}
Critique quality and critique adherence are scored by prompted \texttt{gpt-oss-120B}
judges. Critique quality uses a rubric-for-critiques prompt and observes only
$(x,y_0,c)$. Critique adherence uses a rubric-for-refinement prompt and observes
$(x,y_0,c,y_1)$. Gain is computed as $S(y_1)-S(y_0)$, where $S$ is the official
WritingBench evaluator score. To make the metrics comparable, we report all scalar scores on a common 10-point scale.

\paragraph{Delta computation.}
Table~\ref{tab:motivation-details} reports mean deltas over the 1,000 examples. In the
critic-scaling block, deltas are larger-critic minus smaller-critic with the
actor fixed. In the actor-scaling block, deltas are larger-actor minus
smaller-actor with the critique fixed.

\paragraph{Prompts.}
Tables~\ref{tab:critique-prompt}, \ref{tab:revision-prompt},
\ref{tab:rfc-prompt}, and \ref{tab:refinement-prompt} show the prompts used for
critique generation, revision generation, critique-quality judging, and critique adherence judging.

\begin{table*}[b]
\centering
\small
\setlength{\tabcolsep}{5pt}
\renewcommand{\arraystretch}{1.05}

\begin{tabular}{llccc}
\toprule
Setting & Scaling 
& $\Delta$ Critic\ quality 
& $\Delta$ Critique\ adherence 
& $\Delta$ Gain \\
\midrule
\multicolumn{5}{l}{\textbf{\textit{Critic scaling with the same self-refiner}}} \\
Qwen3-4B  & 8B $\to$ 32B & +0.165 & \textbf{$-$0.169} & +0.060 \\
Qwen3-8B  & 8B $\to$ 32B & +0.132 & \textbf{$-$0.223} & +0.094 \\
Qwen3-32B & 8B $\to$ 32B & +0.102 & +0.014 & +0.017 \\
Llama-3.2-3B  & 8B $\to$ 70B & +0.320 & +0.039 & +0.134 \\
Llama-3.1-8B  & 8B $\to$ 70B & +0.331 & +0.005 & +0.146 \\
\midrule
\multicolumn{5}{l}{\textbf{\textit{Refiner scaling with the same initial response and critique}}} \\
Qwen3-4B $y_0$ + Qwen3-32B critique & 4B $\to$ 32B & fixed & +1.296 & +0.375 \\
Qwen3-8B $y_0$ + Qwen3-32B critique & 8B $\to$ 32B & fixed & +1.001 & +0.314 \\
Llama-3.2-3B $y_0$ + Llama-3.3-70B critique & 3B $\to$ 70B & fixed & +1.759 & +0.561 \\
Llama-3.1-8B $y_0$ + Llama-3.3-70B critique & 8B $\to$ 70B & fixed & +1.094 & +0.322 \\
\bottomrule
\end{tabular}

\caption{
% Controlled zero-shot analysis. Scaling the critic improves standalone critique quality but does not reliably improve uptake or gain for the same self-refiner; in contrast, scaling the refiner with the same critique substantially improves both. All deltas are larger-model minus smaller-model conditions.
Individual comparisons for the controlled zero-shot analysis summarized in
Figure~\ref{fig:motivation}. 
\emph{Critic quality}, \emph{Critique Adherence}, and \emph{Gain} respectively denote
standalone critique quality, how well the actor incorporates the critique into the revision, and downstream
improvement $S(y_1)-S(y_0)$ under the WritingBench evaluator. All scalar scores on a common 10-point scale. The first block
scales only the critic while fixing the self-refining actor; the second block
scales only the refiner while fixing $y_0$ and the critique. All deltas are
larger-model minus smaller-model conditions. Bold indicates decreases in critique adherence.
}
\label{tab:motivation-details}
\end{table*}

% =========================
% Critique generation prompt
% =========================
\begin{table*}[h]
\centering
\small
\setlength{\fboxsep}{7pt}
\setlength{\fboxrule}{0.7pt}

\begin{adjustbox}{max width=\textwidth, max totalheight=0.97\textheight, center}
\fbox{%
\begin{minipage}{0.97\textwidth}

You are a helpful writing critic.

\vspace{0.5em}
Given a writing prompt and a response, provide concise feedback that can help revise the response into a better answer.

\vspace{0.5em}
Focus on the most important issues. Be specific and actionable. Do not rewrite the response.

\vspace{0.8em}
\textbf{Prompt:}

\texttt{\{prompt\}}

\vspace{0.8em}
\textbf{Response:}

\texttt{\{response\}}

\vspace{0.8em}
\textbf{Your Critique:}

\end{minipage}%
}
\end{adjustbox}

\vspace{0.3em}
\caption{Prompt for critique generation.}
\label{tab:critique-prompt}
\end{table*}

% =========================
% Revision generation prompt
% =========================
\begin{table*}[t]
\centering
\small
\setlength{\fboxsep}{7pt}
\setlength{\fboxrule}{0.7pt}

\begin{adjustbox}{max width=\textwidth, max totalheight=0.97\textheight, center}
\fbox{%
\begin{minipage}{0.97\textwidth}

You are a helpful writing assistant.

\vspace{0.5em}
Given a writing prompt, a previous response, and critique feedback, revise the response into a better answer.

\vspace{0.5em}
Use the critique as guidance, but keep the final response natural, coherent, and faithful to the original prompt.
Return the full revised response, not only the changed parts.
Do not explain your changes.

\vspace{0.8em}
\textbf{Prompt:}

\texttt{\{prompt\}}

\vspace{0.8em}
\textbf{Previous Response:}

\texttt{\{response\}}

\vspace{0.8em}
\textbf{Critique:}

\texttt{\{critique\}}

\vspace{0.8em}
\textbf{Revised Response:}

\end{minipage}%
}
\end{adjustbox}

\vspace{0.3em}
\caption{Prompt for revision generation.}
\label{tab:revision-prompt}
\end{table*}

% =========================
% Critique-quality judging prompt
% =========================
\begin{table*}[t]
\centering
\small
\setlength{\fboxsep}{7pt}
\setlength{\fboxrule}{0.7pt}

\begin{adjustbox}{max width=\textwidth, max totalheight=0.97\textheight, center}
\fbox{%
\begin{minipage}{0.97\textwidth}

You are an expert evaluator of writing critiques.

\vspace{0.35em}
Your task is to evaluate the quality of a critique as an intervention for improving an initial response.

\vspace{0.35em}
You will be given:
\begin{enumerate}[leftmargin=1.5em, itemsep=0.05em, topsep=0.1em]
    \item A user prompt
    \item An initial response
    \item A critique of that initial response
\end{enumerate}

\vspace{0.25em}
Evaluate the critique using the rubric below.

\vspace{0.45em}
\textbf{Rubric dimensions.}
\begin{description}[leftmargin=2.0em, itemsep=0.22em, topsep=0.12em]
    \item[\textbf{1. Specificity}]
    Does the critique identify concrete weaknesses or revision targets?
    Low score: vague comments like ``be clearer'' or ``improve flow''.
    High score: points to specific missing content, structural issues, stylistic inconsistencies, factual problems, or weak reasoning.

    \item[\textbf{2. Actionability}]
    Does the critique provide guidance that can be directly acted on in revision?
    Low score: only states that something is bad.
    High score: suggests what should be changed, added, removed, reordered, clarified, or rewritten.

    \item[\textbf{3. Priority}]
    Does the critique focus on the most important issues affecting response quality?
    Low score: focuses on minor surface details while ignoring major flaws.
    High score: emphasizes the most impactful improvements first.

    \item[\textbf{4. Faithfulness}]
    Is the critique grounded in the actual prompt and initial response?
    Low score: invents problems, misreads the response, or asks for changes inconsistent with the prompt.
    High score: accurately reflects what the response did and did not do.

    \item[\textbf{5. Coverage}]
    Does the critique capture the main weaknesses relevant to improving the response?
    Low score: misses major weaknesses.
    High score: addresses the major issues that matter most for improvement.

    \item[\textbf{6. Constructiveness}]
    Is the critique framed in a way that supports effective revision?
    Low score: purely judgmental, dismissive, or unhelpful.
    High score: identifies problems in a way that facilitates revision.
\end{description}

\vspace{0.3em}
\textbf{Scoring.}
For each dimension, assign a score from 1 to 10:
1--2 = very poor, 3--4 = poor, 5--6 = mixed/adequate,
7--8 = good, and 9--10 = excellent.

\vspace{0.3em}
Then provide a short justification for each score from 1 to 10, and a short summary of whether this critique is likely to help a refinement model revise the response.

\vspace{0.3em}
\textbf{Important.}
Evaluate the critique itself, not the absolute quality of the initial response.
Focus on whether this critique is a useful intervention for revision.

\vspace{0.45em}
\textbf{Output format.}
Remove symbols that interfere with JSON parsing, and do not use quotation marks
inside reasons. Return only the following JSON format and nothing else:

\begin{quote}
\ttfamily\footnotesize
\{\\
\hspace*{1em}"specificity": \{"score": ..., "reason": "..." \},\\
\hspace*{1em}"actionability": \{"score": ..., "reason": "..." \},\\
\hspace*{1em}"priority": \{"score": ..., "reason": "..." \},\\
\hspace*{1em}"faithfulness": \{"score": ..., "reason": "..." \},\\
\hspace*{1em}"coverage": \{"score": ..., "reason": "..." \},\\
\hspace*{1em}"constructiveness": \{"score": ..., "reason": "..." \}\\
\}
\end{quote}

\end{minipage}%
}
\end{adjustbox}

\vspace{0.3em}
\caption{Rubric-for-Critiques Prompt.}
\label{tab:rfc-prompt}
\end{table*}

% =========================
% Uptake judging prompt
% =========================
\begin{table*}[t]
\centering
\small
\setlength{\fboxsep}{7pt}
\setlength{\fboxrule}{0.7pt}

\begin{adjustbox}{max width=\textwidth, max totalheight=0.97\textheight, center}
\fbox{%
\begin{minipage}{0.97\textwidth}

You are an expert evaluator of response refinement.

\vspace{0.35em}
Your task is to evaluate how well a refined response follows and implements the critique of an initial response.

\vspace{0.35em}
You will be given:
\begin{enumerate}[leftmargin=1.5em, itemsep=0.05em, topsep=0.1em]
    \item A user prompt
    \item An initial response
    \item A critique of the initial response
    \item A refined response
\end{enumerate}

\vspace{0.25em}
Evaluate the refined response only in terms of critique adherence and implementation.
Do \textbf{not} evaluate the absolute quality of the refined response except as needed to judge whether the critique was followed.

\vspace{0.45em}
\textbf{Rubric dimensions.}
\begin{description}[leftmargin=2.0em, itemsep=0.22em, topsep=0.12em]
    \item[\textbf{1. Main-Issue Adherence}]
    Does the refined response address the main revision requests in the critique?
    Low score: ignores most major critique points.
    High score: directly addresses the key requested revisions.

    \item[\textbf{2. Completeness of Adoption}]
    Were the important critique points fully incorporated rather than only partially or superficially addressed?
    Low score: only limited or partial incorporation.
    High score: the major critique points were comprehensively incorporated.

    \item[\textbf{3. Faithful Implementation}]
    Does the refined response implement the critique in a way that is consistent with the critique's intended meaning?
    Low score: misinterprets, distorts, or incorrectly applies the critique.
    High score: correctly reflects the intended revisions suggested by the critique.

    \item[\textbf{4. Non-Superficial Revision}]
    Does the refinement make substantive changes corresponding to the critique rather than only minor wording edits?
    Low score: mostly cosmetic edits with little real uptake of critique.
    High score: substantive revision that clearly reflects the critique.
\end{description}

\vspace{0.3em}
\textbf{Scoring.}
For each dimension, assign a score from 1 to 10:
1--2 = very poor, 3--4 = poor, 5--6 = mixed/adequate,
7--8 = good, and 9--10 = excellent.

\vspace{0.3em}
Then provide a short justification for each score from 1 to 10, and a short summary of how well the refinement incorporated the critique.

\vspace{0.3em}
\textbf{Important.}
Focus only on critique adherence and implementation.
Do not score overall writing quality, creativity, or task success except insofar as they reveal whether the critique was followed.
A refined response may adhere well to the critique even if its overall quality is not strong.

\vspace{0.45em}
\textbf{Output format.}
Remove symbols that interfere with JSON parsing, and do not use quotation marks
inside reasons. Return only the following JSON format and nothing else:

\begin{quote}
\ttfamily\footnotesize
\{\\
\hspace*{1em}"main\_issue\_adherence": \{"score": ..., "reason": "..." \},\\
\hspace*{1em}"completeness\_of\_adoption": \{"score": ..., "reason": "..." \},\\
\hspace*{1em}"faithful\_implementation": \{"score": ..., "reason": "..." \},\\
\hspace*{1em}"non\_superficial\_revision": \{"score": ..., "reason": "..." \}\\
\}
\end{quote}

\end{minipage}%
}
\end{adjustbox}

\vspace{0.3em}
\caption{Rubric-for-Refinement Prompt.}
\label{tab:refinement-prompt}
\end{table*}

\section{Qualitative Analysis of Critic--Actor Reachability}
\label{app:critic-reachability}

\paragraph{Representative case.}
A paired rollout illustrates one way in which higher standalone critique quality can coincide with lower critique adherence. For a martial-arts story
prompt, the \texttt{Qwen3-32B} critique receives a higher standalone
quality score than the \texttt{Qwen3-8B} critique (9.00 vs.\ 8.33), but
yields substantially lower adherence from the same \texttt{Qwen3-8B}
refiner (3.50 vs.\ 8.75). The 8B critic requests localized changes, such as
introducing a rival and strengthening the mentor's role, which the refiner
largely implements. The 32B critic instead proposes a coordinated redesign
spanning the story's symbolism, character motivations, climax, ending, and
dialogue, of which the refiner adopts only isolated elements. This
illustrates an actor-relative \emph{reachability gap}: feedback that
appears stronger in isolation may be less useful when its requested changes
exceed the target actor's revision capability.

\paragraph{Case selection.}
We examine paired rollouts from the critic-scaling condition with the
instruction, initial response, and \texttt{Qwen3-8B} refiner fixed. We select
English-language cases in which the \texttt{Qwen3-32B} critic receives a higher
standalone critique-quality score than the \texttt{Qwen3-8B} critic but produces lower
critique adherence. We report four cases from distinct writing
tasks to make the corresponding revision behavior legible.

\paragraph{Paired cases.}
Table~\ref{tab:critic-reachability-cases} compares the main-text example with
three additional cases. For each critic, we distinguish the requested changes
from those observed in the corresponding revision.

\begin{table*}[t]
\centering
\footnotesize
\setlength{\tabcolsep}{3.5pt}
\renewcommand{\arraystretch}{1.04}
\begin{tabular}{p{0.17\linewidth}p{0.36\linewidth}p{0.36\linewidth}}
\toprule
Prompt and scores & \texttt{Qwen3-8B} critique and revision & \texttt{Qwen3-32B} critique and revision \\
\midrule

\textbf{Martial-arts story} (main-text example)

Revise a story so that calligraphy shapes its characters, conflict, and themes.

\textit{Critique Quality:} 8.33 $\to$ 9.00

\textit{Critique Adherence:} 8.75 $\to$ 3.50
&
\textbf{Requested:} Add a former rival or confidant, connect calligraphy to
freedom, and make the ending show the journey's social impact.

\textbf{Observed:} The revision adds a former rival, expands the mentor, and
shows villagers adopting the protagonist's calligraphy as a symbol of peace.
&
\textbf{Requested:} Coordinate the calligraphic symbolism, antagonist motives,
rival arc, climax, collective ending, and dialogue.

\textbf{Observed:} The revision adopts isolated symbols and one subtler
exchange, but does not connect the rival to the climax or make the ending
collective, and retains didactic dialogue. \\
\midrule

\textbf{Quantum-computing report}

Write an academic report on quantum computing for engineering optimization,
including recent research and technical details.

\textit{Critique Quality:} 8.67 $\to$ 8.83

\textit{Critique Adherence:} 8.00 $\to$ 2.50
&
\textbf{Requested:} Add a formal chapter structure, recent references, concrete
engineering case studies, deeper QAOA/VQE discussion, hybrid methods, and
implementation challenges.

\textbf{Observed:} The revision adds chapters, references, case studies,
QAOA/VQE, hybrid methods, and implementation challenges.
&
\textbf{Requested:} In addition to restructuring, add named recent studies,
gate-level mathematical formulations, Qiskit/Cirq details, hybrid workflows,
and ethical and regulatory analysis.

\textbf{Observed:} The revision adds technical depth and named references but
omits Qiskit/Cirq and ethics/regulation; its exact citation claims also remain
weakly grounded. \\
\midrule

\textbf{Consumer-product comparison}

Compare AirPods Pro 2 and Sony WF-1000XM5 using concrete evidence about design,
sound, noise cancellation, battery, and user experience.

\textit{Critique Quality:} 8.50 $\to$ 8.67

\textit{Critique Adherence:} 7.00 $\to$ 2.00
&
\textbf{Requested:} Improve the comparison structure, add missing metrics,
explain technical terms, and tailor the conclusion to different user needs.

\textbf{Observed:} The revision adds comparative sections, definitions,
additional metrics, and a user-oriented conclusion.
&
\textbf{Requested:} Quantify quick-charge differences, connect each model's
noise-cancellation strengths to specific use cases, add driver-level details,
and make the pros/cons consistent with the cited evidence.

\textbf{Observed:} The revision mentions several requested features but does
not consistently quantify charging, ground noise-cancellation claims in the
requested use cases, provide the driver comparison, or revise the generic
pros/cons. \\
\midrule

\textbf{Birthday love poem}

Revise a 12--16-line modern love poem with gentle language and natural imagery.

\textit{Critique Quality:} 7.83 $\to$ 8.67

\textit{Critique Adherence:} 8.25 $\to$ 3.00
&
\textbf{Requested:} Tighten wordy lines, replace generic river and sky
metaphors, add subtle emotional tension, and echo earlier imagery at the end.

\textbf{Observed:} The revision directly implements these localized changes,
including ``river's whisper,'' the cold/warm contrast, and a final image that
returns to the lily and stream.
&
\textbf{Requested:} Redesign the poem with varied line lengths and enjambment,
12--14 lines, concrete sensory images, a seasonal emotional arc, and a quiet
ending.

\textbf{Observed:} The revision borrows several suggested images but grows to
17 lines, retains multiple flagged clich\'es, does not develop the requested
seasonal progression, and ends with another cosmic declaration. \\

\bottomrule
\end{tabular}
\caption{Four paired English-language cases in which critic scaling increases standalone
critique quality but decreases adherence from the same \texttt{Qwen3-8B} refiner.
Arrows report \texttt{Qwen3-8B}-critic $\to$ \texttt{Qwen3-32B}-critic scores.}
\label{tab:critic-reachability-cases}
\end{table*}

Across these cases, the revisions produced from the larger critic do not ignore the feedback outright. Rather, they implement isolated suggestions while failing on constraints that require multiple changes to be coordinated or on requests that depend on information unavailable in the revision context. The cases therefore illustrate a mismatch between standalone critique quality and actor-executable guidance: a critique can be more comprehensive yet less reachable for a fixed refiner. They should not be interpreted as evidence that smaller critics are generally superior.
% \clearpage

\section{Query Selection and Sampling}
\label{app:query_selection}

For DeepWriting-20K, we first applied lightweight rule-based filters to remove non-writing examples, overly short prompts, and image-generation prompts. The
remaining examples were scored by an LLM judge (\texttt{gpt-oss-120B}), using the query-filtering prompt shown in Table~\ref{tab:deepwriting_filtering_prompt}. We retained all examples with scores of 4 or 5, yielding 4,937 high-quality writing prompts. Since this was
below our target size of 9,000, we supplemented this set with 4,063 randomly sampled score-3 examples. The resulting set was shuffled and split into three shards of $K{=}3{,}000$ examples each.

For OpenScholar, we used an LLM-based query-quality filtering pipeline following the query-filtering prompt from DR-Tulu~\citep{shao2025dr}. Starting from 55,774
OpenScholar queries, we removed examples with unsupported intents, such as Other, Metadata, or Cannot determine, as well as very short or non-English
queries. We then scored the remaining queries with \texttt{gpt-oss-120B}, and selected all queries scored 4 or
5, producing 17,022 high-quality queries. We sampled 9,000 queries from this high-quality pool and split the result into three shards of $K{=}3{,}000$ queries each.

\begin{table*}[t]
\centering
\footnotesize
\setlength{\fboxsep}{6pt}
\setlength{\fboxrule}{0.7pt}

\begin{adjustbox}{max width=\textwidth, max totalheight=0.86\textheight, center}
\fbox{%
\begin{minipage}{0.96\textwidth}

\textbf{Task.}
You are a task-quality grader for long-form writing-model training.
Grade each DeepWriting query for writing task quality.

\vspace{0.35em}
\textbf{Input fields.}
Use the provided fields:
\begin{itemize}[leftmargin=1.4em, itemsep=0.05em, topsep=0.1em]
    \item \texttt{query}: the user writing instruction
    \item \texttt{ability}: dataset ability label when available
    \item \texttt{domain}: broad writing domain when available
    \item \texttt{category}: detailed writing category when available
    \item \texttt{language\_hint}: language hint when available
\end{itemize}

\vspace{0.25em}
\textbf{Skip criteria.}
Skip a query if it is not a writing task, is unsafe, asks for private/PII harvesting,
is image-only, is pure math/coding/QA, is empty/trivial, or is only
translation/rewrite/formatting.

\vspace{0.35em}
\textbf{Scoring rubric.}
Use integers only:
\begin{description}[leftmargin=2.0em, style=nextline, itemsep=0.1em, topsep=0.1em]
    \item[\textbf{1}] Not writing, unsafe, image-only, empty/trivial, pure QA/math/code, or not evaluable.
    \item[\textbf{2}] Weak writing task: generic, boilerplate, mostly rewriting/formatting, or too little composition.
    \item[\textbf{3}] Normal valid writing task; broad default for ordinary essays, articles, stories, reports, speeches, emails, lesson plans, reviews, and marketing copy.
    \item[\textbf{4}] Strong writing task requiring meaningful planning around audience, structure, evidence, tone, domain framing, scenario logic, or provided materials.
    \item[\textbf{5}] Excellent deep-writing task requiring a coherent, domain-specific artifact that integrates materials, data, case facts, evidence, stakeholder goals, or multiple writing criteria. Length or many fields are not enough.
\end{description}

\vspace{0.35em}
\textbf{Strict calibration.}
\begin{itemize}[leftmargin=1.4em, itemsep=0.05em, topsep=0.1em]
    \item Most valid DeepWriting tasks should receive 3, not 4.
    \item Do not give 4 just because the query is long, sectioned, domain-specific, or creative.
    \item Use 4 only when constraints materially change the writing plan. If unsure between 3 and 4, choose 3.
    \item Use 5 when output quality depends on integrating materials and constraints, not merely following a format.
\end{itemize}

\vspace{0.35em}
\textbf{Calibration examples.}
\begin{itemize}[leftmargin=1.4em, itemsep=0.05em, topsep=0.1em]
    \item \texttt{"hello"} $\rightarrow$ 1
    \item \texttt{"Draw a poster for a coffee shop"} $\rightarrow$ 1
    \item \texttt{"Write a professional article about balcony gardening."} $\rightarrow$ 3
    \item \texttt{"Write a lesson plan on ecosystems with objectives, activities, and assessment."} $\rightarrow$ 3
    \item \texttt{"Write a grant proposal summary for a community clinic using budget constraints and evaluation metrics."} $\rightarrow$ 4
    \item \texttt{"Write a launch memo for a budgeting app balancing user needs, compliance risks, and market positioning."} $\rightarrow$ 4
    \item \texttt{"Using provided case facts, write a legal judgment section explaining custody and property division."} $\rightarrow$ 5
    \item \texttt{"Write a policy brief reconciling conflicting evidence for expert and public stakeholders."} $\rightarrow$ 5
    \item \texttt{"Write a paper outline using six months of HVAC data and DRL optimization records for a journal submission."} $\rightarrow$ 5
\end{itemize}

\vspace{0.35em}
\textbf{Output format.}
Return JSON only in this exact shape:
\begin{quote}
\ttfamily\scriptsize
\{\\
\hspace*{1em}"results": [\\
\hspace*{2em}\{\\
\hspace*{3em}"id": "<same id as input>",\\
\hspace*{3em}"score": <integer 1--5>,\\
\hspace*{3em}"decision": "SKIP" or "GRADE",\\
\hspace*{3em}"rationale": "<brief reason, max 25 words>"\\
\hspace*{2em}\}\\
\hspace*{1em}]\\
\}
\end{quote}

\vspace{-0.2em}
\textbf{Decision rule.}
Use \texttt{SKIP} for scores 1--2 and \texttt{GRADE} for scores 3--5.

\end{minipage}%
}
\end{adjustbox}

\vspace{0.3em}
\caption{Prompt for filtering DeepWriting-20K queries.}
\label{tab:deepwriting_filtering_prompt}
\end{table*}

\section{Additional Experimental Details}
\label{app:exp_details}

\subsection{Cross-Model Agreement of the Reward Judge}
\label{app:judge-validation}

Our critic-training reward is computed by \texttt{gpt-oss-120B}. To assess
whether candidate rankings remain consistent when changing the reward judge,
we compare its judgments with those of \texttt{Claude Opus 4.8}, a model from
a different provider.

We randomly sample 200 prompts from DeepWriting-20K and generate four
critique--revision rollouts per prompt using \texttt{Qwen3-8B} as both the
critic and the actor, yielding 800 rollouts in total. Both judges
independently score the same rollouts using the identical \score rubric and
judging prompt, without access to each other's scores. Claude is used only for
this analysis and does not contribute to critique generation, revision
generation, or training.

Because \score is discrete, multiple candidates can receive the same score.
For each prompt and judge, we treat all highest-scoring candidates as the
judge's top set. Top-1 overlap is the percentage of prompts for which the two
judges' top sets share at least one candidate. For pairwise agreement, we consider all six pairwise comparisons among the
four candidates generated for each prompt. We retain only comparisons for
which both judges express a strict preference and measure how often they
prefer the same candidate. Comparisons tied by either judge are excluded.

The judges achieve a top-1 overlap of 89.5\% and agree on 77.9\% of candidate
pairs for which both express a strict preference. These results indicate that
the candidate rankings used to assign relative credit during GRPO are largely
consistent across the two reward judges.

\subsection{Training Hyperparameters}
\label{app:hyperparams}

Table~\ref{tab:hyperparams} summarizes the core hyperparameters used for
critic GRPO training and actor DPO training.

\subsection{Actor-Training Pair Construction}
\label{app:pair-construction}

We construct three shared candidate-query pools, each containing
$K{=}3{,}000$ unique queries. The pools are mutually disjoint. For each
actor-training condition and each query in a pool, we generate one candidate
rollout $(x,y_0,c,y_1)$, where $y_1$ is the revision produced from critique
$c$. Thus, within a condition, each query can contribute at most one
actor-training pair.

We compare each revised response $y_1$ with its corresponding initial response
$y_0$ using a response-level pairwise assessment. The
\texttt{gpt-oss-120B} judge receives the instruction and the two responses in
randomized order, without access to the critique. Candidates for which $y_1$ is
preferred over $y_0$ are eligible for actor training; ties and preferences for
$y_0$ are excluded.

For each condition and candidate-query pool, we uniformly sample 2,000 eligible
pairs without replacement and define $(y^+,y^-)=(y_1,y_0)$. Co-evolution uses
the 2,000 pairs sampled from each pool for the corresponding actor update.
Single-stage conditions concatenate the pairs sampled from the three pools and
perform one actor update on the resulting 6,000 pairs. Across conditions, we
hold fixed the underlying candidate queries, the number of candidates generated
per query, the pairwise assessment and subsampling procedures, and the number
of DPO training pairs.

\subsection{Critic-Training Rewards and Judge Prompts}
\label{app:reward_prompts}

We use three reward definitions for critic training.
The \emph{critique-quality reward} scores the critique $c$ in isolation given
$(x,y_0)$, using a rubric-for-critiques prompt that evaluates whether the
feedback is specific, actionable, important, faithful, comprehensive, and
constructive. We compute
the raw critique-quality reward as the average of the six criterion scores.

The \emph{outcome-gain reward} scores whether the revised response $y_1$
improves over the initial response $y_0$ for the instruction $x$, without
observing the critique $c$. The judge outputs one of three values:
$1$ if $y_1$ is better than $y_0$, $0.5$ if they are roughly equal, and $0$ if
$y_1$ is worse.

Our \score reward observes the full tuple $(x,y_0,c,y_1)$ and uses
a final scalar score that jointly evaluates the critique and the resulting
revision. This score reflects
whether the critique targets a real weakness, whether the actor incorporates
the feedback, whether the targeted aspect improves, and whether the critique
and revision remain faithful to the original instruction.

For GRPO training, each reward definition is used in a separate condition. Raw
reward values may have different scales across conditions, but they are
normalized within each sampled critique group before computing advantages.
The exact prompt templates for the critique-quality reward, outcome-gain
reward, and \score reward are shown in
Tables~\ref{tab:rfc-prompt}, \ref{tab:outcome_gain_prompt}, and
\ref{tab:taiscore_prompt}, respectively.

\subsection{Training Details}
Unless otherwise noted, we use the same hyperparameters across all training
runs. Table~\ref{tab:hyperparams} summarizes the critic GRPO, actor DPO, and
generation settings. The actor/refiner checkpoint is updated across
co-evolution rounds, while the judge is kept fixed.

\subsection{Critique-Content Controls}
\label{app:critique-content-controls}

To distinguish the effect of critique content from generic second-pass
revision, we compare four revision conditions. All conditions use the same
WritingBench prompts, initial responses $y_0$, \texttt{Qwen3-8B} reviser,
revision prompt, decoding configuration, and evaluator. They differ only in
the feedback supplied to the reviser. We include every generated revision
without preference filtering.

\paragraph{No critique.}
The reviser receives the instruction and initial response and is asked to
improve the response without receiving any critique.

\paragraph{Generic critique.}
The reviser receives a fixed, prompt-agnostic critique that asks it to improve
the response along general dimensions such as correctness, clarity, relevance,
and overall quality. The same critique is used for every example and does not
identify any instance-specific issue or suggest a concrete revision.

\paragraph{Shuffled critique.}
We randomly permute the critiques produced by the \score-trained critic across
examples, ensuring that no response receives its own critique. Thus, this
condition preserves the critic source, critique style, and marginal critique
distribution while breaking the correspondence between each critique and its
instruction--response pair.

\paragraph{\score critique.}
Each initial response receives the critique generated for that same example by
the \score-trained critic. This is the only condition in which the feedback is
both model-generated and correctly matched to the instance.

The no-critique and generic-critique conditions measure improvement attributable
to an additional revision pass and a general request for improvement,
respectively. The shuffled condition further controls for the presence and
form of \score-generated feedback while removing its instance-specific
relevance.

\begin{table*}[t]
\centering
\small
\setlength{\tabcolsep}{5pt}
\renewcommand{\arraystretch}{1.3}
\begin{tabular}{llll}
\toprule
\multicolumn{2}{c}{\textbf{Critic GRPO / Judging}} &
\multicolumn{2}{c}{\textbf{Actor DPO / Generation}} \\
\cmidrule(r){1-2}\cmidrule(l){3-4}
\textbf{Component} & \textbf{Setting} &
\textbf{Component} & \textbf{Setting} \\
\midrule
Critic model & \texttt{Qwen3-8B}
& Actor/refiner model & Current actor checkpoint \\
LLM judge & \texttt{gpt-oss-120B}
& GPU infrastructure & $4\times$ NVIDIA RTX PRO 6000 Blackwell \\
Critic training algorithm & GRPO
& Actor training algorithm & DPO \\
Critiques per prompt & $N=4$
& DPO $\beta$ & 0.1 \\
Critic learning rate & $1\times 10^{-6}$
& DPO learning rate & $1\times 10^{-6}$ \\
KL coefficient & 0.02
& DPO epochs & 1 \\
Critic train batch size & 8
& DPO max length & 4096 \\
GRPO mini-batch size & 8
& Max prompt length & 2048 \\
Rollout temperature / top-$p$ & 0.8 / 0.95
& Max response length & 4096 \\
Judge temperature / top-$p$ & 0.2 / 1.0
& Refiner temperature / top-$p$ & 0.2 / 0.95 \\
\bottomrule
\end{tabular}
\caption{Core hyperparameters used for critic GRPO and actor DPO training.}
\label{tab:hyperparams}
\end{table*}

\begin{table*}[t]
\centering
\small
\setlength{\fboxsep}{6pt}
\begin{adjustbox}{width=\textwidth}
\fbox{%
\begin{minipage}{0.96\textwidth}

\textbf{Instruction.}
You are an expert evaluator for open-ended writing tasks.

\vspace{0.45em}
\textbf{Task.}
Given a user prompt and two responses, decide whether Response B is better
than Response A overall, considering writing quality, relevance, clarity,
completeness, fluency, and faithfulness without favoring unnecessary length
or irrelevant elaboration.

\vspace{0.45em}
\textbf{Input.}
You will be given:
\begin{itemize}[leftmargin=2.0em, itemsep=0.15em, topsep=0.12em]
    \item User prompt
    \item Response A: initial response
    \item Response B: revised response
\end{itemize}

\vspace{0.45em}
\textbf{Output format.}
Return only one token as the final answer:
\begin{description}[leftmargin=2.0em, itemsep=0.18em, topsep=0.12em]
    \item[\textbf{1}]
    if Response B is better than Response A

    \item[\textbf{0.5}]
    if they are roughly equal

    \item[\textbf{0}]
    if Response B is worse than Response A
\end{description}

\vspace{0.45em}
\textbf{Important.}
Do not return JSON.
Do not return any explanation.
Do not return any other text.

\end{minipage}%
}
\end{adjustbox}

\vspace{0.3em}
\caption{Outcome-gain reward prompt used to compute the baseline training reward.}
\label{tab:outcome_gain_prompt}
\end{table*}

\begin{table*}[t]
\centering
\small
\setlength{\fboxsep}{6pt}
\begin{adjustbox}{width=\textwidth}
\fbox{%
\begin{minipage}{0.96\textwidth}

\textbf{Instruction.}
You are evaluating a writing critique as a training reward.

\vspace{0.45em}
\textbf{Input.}
You will be given:
\begin{itemize}[leftmargin=2.0em, itemsep=0.15em, topsep=0.12em]
    \item User prompt
    \item Response A: initial response
    \item Critique of Response A
    \item Response B: revised response after the critique
\end{itemize}

\vspace{0.3em}
Judge whether the critique was a useful intervention.

\vspace{0.45em}
\textbf{Rubric dimensions.}
Score these dimensions from 1 to 10. All scores must be integers.

\begin{description}[leftmargin=2.0em, itemsep=0.22em, topsep=0.12em]
    \item[\textbf{1. Critique Quality}]
    Is the critique faithful, specific, important, and actionable?

    \item[\textbf{2. Critique Uptake}]
    Does Response B actually follow the critique?

    \item[\textbf{3. Targeted Quality Gain}]
    Does Response B improve over Response A on the issue targeted by the critique?

    \item[\textbf{4. Prompt Faithfulness}]
    Do the critique and Response B stay aligned with the user prompt?
\end{description}

\vspace{0.45em}
\textbf{Final score.}
Give a final score, also an integer from 1 to 10.

\vspace{0.3em}
Give high scores only when the critique identifies important, fixable issues
and Response B improves by following it.
Give low scores when the critique is vague, generic, or unfaithful, or when
Response B improves for reasons unrelated to the critique.
Penalize Response B if it becomes worse, overlong, irrelevant, or less faithful
to the prompt.
Do not reward unnecessary length or superficial paraphrasing.

\vspace{0.45em}
\textbf{Output format.}
Return valid JSON only:

\begin{quote}
\ttfamily\footnotesize
\{\\
\hspace*{1em}"critique\_quality": \{"score": ..., "reason": "..." \},\\
\hspace*{1em}"critique\_uptake": \{"score": ..., "reason": "..." \},\\
\hspace*{1em}"quality\_gain": \{"score": ..., "reason": "..." \},\\
\hspace*{1em}"prompt\_faithfulness": \{"score": ..., "reason": "..." \},\\
\hspace*{1em}"score": ...\\
\}
\end{quote}

\end{minipage}%
}
\end{adjustbox}

\vspace{0.3em}
\caption{\score judge prompt used to compute the critique GRPO reward.}
\label{tab:taiscore_prompt}
\end{table*}

\section{Human Evaluation Protocol}
\label{app:human-eval}

We conduct a blind pairwise evaluation of three comparisons central to our
main claims: \textsc{Taiscore} versus the base actor, \textsc{Taiscore} versus
the off-the-shelf \texttt{gpt-oss-120B} critic baseline, and
\textsc{Taiscore} with co-evolution versus single-stage \textsc{Taiscore}. We
randomly sample 50 prompts from WritingBench and evaluate all three comparisons
on these prompts, yielding 150 response pairs and 450 individual judgments.

For each response pair, three annotators independently view the instruction and
two responses. We anonymize the system identities and training conditions,
randomize response order, and do not show annotators automatic evaluation
scores. Annotators select the response that better fulfills the instruction
overall or select a tie when neither response is clearly preferable. We assign
the outcome selected by at least two annotators as the per-prompt majority
decision. If the three annotators respectively select the first response, the
second response, and a tie, we record the item as having no majority. Table
\ref{tab:human-eval} reports wins and losses from the perspective of the first
condition named in each comparison.

% \section{Example Appendix}
% \label{sec:appendix}

% This is an appendix.

\end{document}